\documentclass{article} % For LaTeX2e
\usepackage[final]{colm2026_conference}

\usepackage{microtype}
\usepackage{hyperref}
\usepackage{url}
\usepackage{booktabs}
\usepackage{enumitem}
\usepackage[ruled,vlined]{algorithm2e}
\usepackage{changepage} % for adjustwidth
\usepackage{setspace}   % for setstretch (optional but nice)
\usepackage{multirow}
\usepackage{makecell}
\usepackage{array}
\usepackage{graphicx}
\usepackage{arydshln}
\usepackage{subcaption}
\usepackage{wrapfig} 
\usepackage{amsmath}

\usepackage{lineno}

\definecolor{darkblue}{rgb}{0, 0, 0.5}
\hypersetup{colorlinks=true, citecolor=darkblue, linkcolor=darkblue, urlcolor=darkblue}

\title{Training LLMs with Reinforcement Learning for Intent-Aware Personalized Question Answering}

\title{Training LLMs with Reinforcement Learning for Intent-Aware Personalized Question Answering}

\title{Training LLMs with Reinforcement Learning for Intent-Aware Personalized Question Answering}

\author{
\textbf{
Maryam Amirizaniani$^{1}$, Benjamin Charles Germain Lee$^{1}$, Jevin West$^{1}$, Nicholas Weber$^{1}$
} \\
$^{1}$University of Washington, Seattle, WA, USA \\
\texttt{\{amaryam, bcgl, jevinw, nmweber\}@uw.edu}
}

\begin{document}
\maketitle

\begin{abstract}
Effective personalized question answering (PQA) in language models requires grounding responses in the user's underlying intent, where intent refers to the implicit “why” behind a query beyond its explicit wording. However, existing approaches to intent-aware personalization rely on multi-turn conversational context or rich user profiles, and do not explicitly model user intent during reasoning process. This limits their effectiveness in single-turn settings, where the user's latent goal must be inferred from minimal input and integrated into the thinking and reasoning process. To bridge this gap, we propose IAP (\textbf{I}ntent \textbf{A}ware \textbf{P}ersonalization), a reinforcement learning framework that trains models to infer implicit user intent directly from a single-turn question and incorporate it into thinking steps through a tag-based schema for generating personalized, intent-grounded answers. By optimizing intent-aware answer trajectories under a personalized reward function, IAP reinforces generation paths that make implicit user intent explicit and produce responses that better align with the user's underlying goal. Through experiments on the LaMP-QA benchmark across six models, IAP consistently outperforms all baselines, achieving an average macro-score gain of around 7.5\% over the strongest competitor, demonstrating that modeling implicit user intent within the training objective is a promising direction for PQA.
 
\end{abstract}

\section{Introduction}
Personalized question answering (PQA) aims to generate user-tailored answers and can be further improved by aligning responses with the user’s underlying intent~\citep{zhang-etal-2024-discrimination, zhao2025do}. In this context, intent refers to the underlying goal or purpose behind a user’s query (e.g., seeking emotional support or practical guidance)~\citep{chatterjee-sengupta-2020-intent}. Since intent is often implicit, accurately inferring it is critical for generating answers that are not only correct but also aligned with what the user actually needs~\citep{10.1145/3732294, 10.1145/3627673.3680273}. For example, consider the question, ``Should I change my job?'' Two users can ask the same question with different implicit intents: one may be seeking decision-analytic support (e.g., comparing options and trade-offs), while another may want help interpreting uncertainty and diagnosing their source of dissatisfaction (e.g., burnout, values mismatch, or limited growth). If a system responds to both users with a generic pros and cons list, the answer may still be coherent yet misaligned because it is too abstract, too clinical, or too vague for the user’s intended outcome.

To understand user intent, prior work has explored intent modeling from several angles, ranging from classic closed-domain approaches that assign user utterances to predefined intent categories~\citep{larson-etal-2019-evaluation, casanueva-etal-2020-efficient, 10.1145/3732294} to more recent Large Language Model (LLM)-based methods that use instruction tuning to extract intents~\citep{chatterjee-sengupta-2020-intent, zhao2025do, 10.1145/3627673.3679832, mo-etal-2023-convgqr, mitra2025recap, askari-etal-2025-solid}. Despite this progress, three key limitations remain: 
\begin{figure*}[htbp]
  \centering
  \vspace{-0.4cm}
  \includegraphics[scale=0.25]{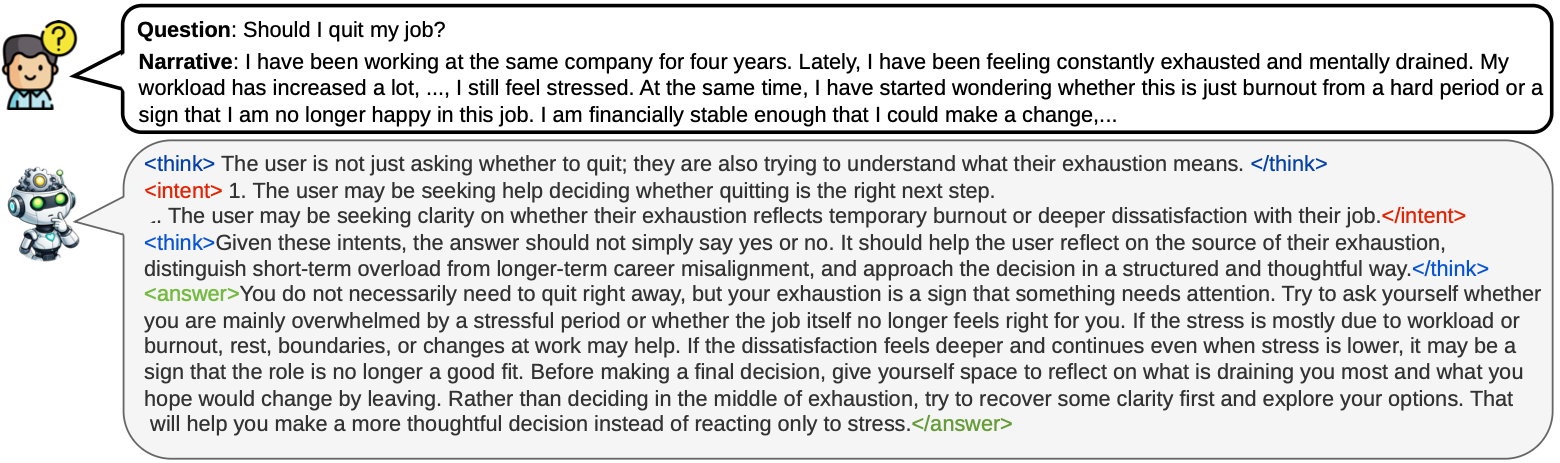}  
   \vspace{-0.2cm}
  \caption{An example of IAP.}
  \label{fig:example}
  \vspace{-0.5cm}
\end{figure*}
(1) current approaches primarily infer intents through prompting, without incorporating intent into the model’s reasoning and training processes before personalized answer generation~\citep{kim2025ipqa, bodonhelyi2024user, 10.1145/3627673.3679832, askari-etal-2025-solid}; (2) most depend on multi-turn dialogue history or detailed user profiles, limiting their applicability in single-turn question answering (QA) settings where multiple user intents must be inferred from a single question alone~\citep{askari-etal-2025-solid, larson-etal-2019-evaluation}; and (3) existing methods optimize LLMs for personalized answer generation using only intent-aware signals, while neglecting the role of intent-free responses~\citep{zhao2026improving, askari-etal-2025-solid}. Building on these gaps, we investigate the following research questions:

\textbf{RQ1:} How can we train a model to infer a user's latent intents and use it to generate personalized answers in single-turn question answering?

\textbf{RQ2:} To what extent is leveraging contrastive intent-aware and intent-free response signals in training LLMs effective for generating user-intent-grounded answers?

To answer these research questions, we propose IAP (\textbf{I}ntent \textbf{A}ware \textbf{P}ersonalization), a Reinforcement Learning (RL) framework that trains LLMs to infer a user's latent intents from a single-turn input question and incorporate it into their thinking process before generating personalized, intent-grounded answers. IAP instantiates this learning objective using Decoupled Clip and Dynamic sAmpling Policy Optimization (DAPO)~\citep{yu2025dapo}, which optimizes intent-aware answer trajectories under a personalized reward function that evaluates how well a response aligns with the user's intent and preferences. The personalized reward further incorporates an auxiliary intent-free response signal that penalizes outputs close to non-contextualized answers, thereby discouraging generic generation and encouraging responses that are more sensitive to user intent and better reflect the user’s preferences. Following~\citep{mo-etal-2023-convgqr, mitra2025recap}, IAP incorporates inferred intent into the model’s thinking and reasoning process through a tag-based format, making the user’s latent goal explicit before answer generation. Specifically, the process is structured into a {\color{blue}\texttt{<think>}} tag for internal reasoning, an {\color{red}\texttt{<intent>}} tag for explicitly capturing the user’s latent intent, and an {\color{teal}\texttt{<answer>}} tag for generating the final personalized response, as illustrated in Figure~\ref{fig:example}. To evaluate IAP, we use the LaMP-QA dataset~\citep{salemi-zamani-2025-lamp}, which consists of long-form social questions, and study six different LLMs to assess whether intent-aware reasoning improves PAQ. Experimental results show that IAP consistently outperforms all baselines, achieving an average macro-score gain of approximately 7.5\% over the strongest competitor across all evaluated LLMs.

%The \textbf{motivation} of this research is to infer implicit intent from minimal text, since users in forums like Reddit and Stack Exchange often ask standalone questions without prior context, unlike the rich dialogue history assumed by existing intent modeling methods. The \textbf{novelty} of this study lies in implicit intent understanding and modeling based on question answering from surface-level intent classification to modeling the user’s implicit “why”. To address this gap, the study makes three main  \textbf{contributions}

The main contributions of this paper are summarized as follows: \textbf{(1)} We introduce IAP, an RL-based framework that trains LLMs to infer implicit user intents from a single-turn question and incorporate them into the reasoning process via a tag-based schema to generate personalized, intent-grounded answers\footnote{The code is available at:~\url{https://github.com/maryamamiri114/IAP}}. \textbf{(2)} We leverage a contrastive reward function that explicitly contrasts intent-aware and intent-free responses during training to discourage generic generation and push the model toward more intent-sensitive outputs, and \textbf{(3)} We evaluate IAP on the LaMP-QA across six LLMs, demonstrating consistent improvements of approximately 7.5\% on average over the strongest baseline across all models.
\begin{figure*}[htbp]
  \centering
  \includegraphics[scale=0.5]{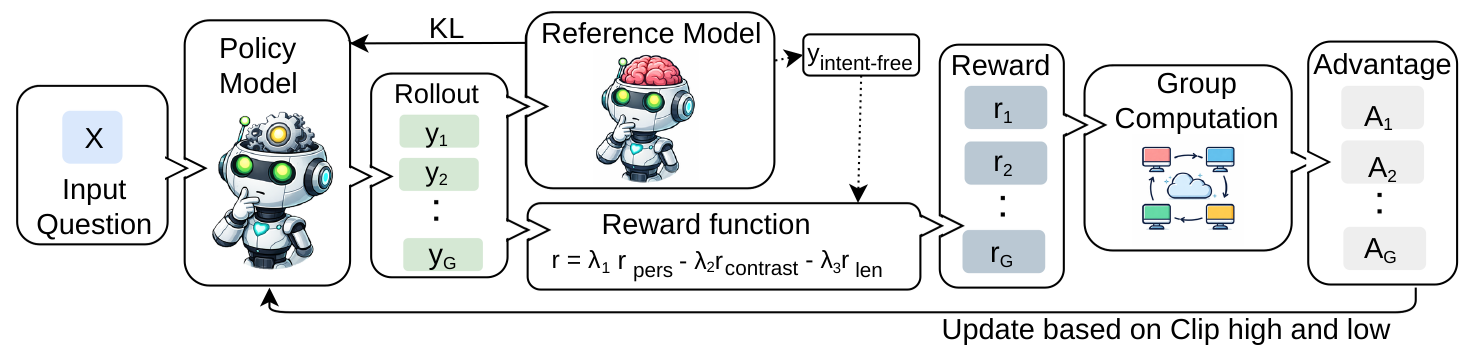}  
  \caption{IAP Framework for intent-aware personalization question answering.}
  \label{fig:framework1}
  \vspace{-0.5cm}
\end{figure*}
\section{Related Work}
%PQA aims to generate user-tailored answers and can be approached in various ways~\citep{salemi-etal-2024-lamp, salemi2025pathways, amirizaniani2026learning, xie2025a}, with intent understanding emerging as one of the most effective~\citep{kim2025ipqa}. Traditional approaches operationalize intent as predefined label categories, including datasets such as BANK77~\citep{casanueva-etal-2020-efficient}, CLINC150~\citep{larson-etal-2019-evaluation}, SNIPS~\citep{coucke2018snips}, MultiWOZ~\citep{budzianowski-etal-2018-multiwoz}, MultiDoGO~\citep{peskov-etal-2019-multi}, and SGD~\citep{rastogi2020towards}. However, these approaches struggle in open-domain settings where intents are ambiguous, compositional, and evolving~\citep{10.1145/3732294}.

Intent understanding is among the most effective strategies for personalized question answering~\citep{kim2025ipqa, salemi-etal-2024-lamp, salemi2025pathways, xie2025a, 10.1145/3731120.3744598, 10.1145/3701551.3703576}. Existing work, however, largely treats intent as classification over predefined label taxonomies, as reflected in benchmarks like BANK77~\citep{casanueva-etal-2020-efficient}, CLINC150~\citep{larson-etal-2019-evaluation}, and SGD~\citep{rastogi2020towards}. These fixed-category formulations struggle in open-domain settings where intents are ambiguous, compositional, and evolving~\citep{10.1145/3732294}

As LLMs have become increasingly prevalent, researchers have begun leveraging them to extract user intent in open-domain settings~\citep{vedula2019towards, anderson-etal-2024-open, ahmad2025multi}, including reformulating intent detection as a language generation problem to enable inference from natural-language descriptions~\citep{zhang-etal-2024-discrimination}. Incorporating conversational history and situational cues has further been shown to resolve ambiguity in underspecified queries, underscoring the value of context-aware intent modeling~\citep{chatterjee-sengupta-2020-intent, zhu-etal-2025-survey, 10.1145/3682064, 10.1016/j.procs.2022.09.444, atuhurra2024domain, zhang2024mintrec, MIntRec}. Building on this, CA-UIA-QG generates follow-up questions aligned with shifting user intent~\citep{Dong2024}, and Chain-of-Intent integrates sequence models with LLM generation to produce intent-guided dialogues~\citep{yin-etal-2025-eclm}. Beyond text, IntentQA further demonstrates that intent reasoning in video settings requires contextual and temporal grounding~\citep{li2023intentqa}.

To further extract implicit intents, prior research has employed various methods. ~\citet{mo-etal-2023-convgqr} combines query rewriting with answer generation to ground implied intent, while ~\citet{mitra2025recap} frames rewriting as capturing a user's most recent goals to reduce intent mismatches. ~\citet{zhao2026improving} extracts user intent through writing and citation intents, and ~\citet{askari-etal-2025-solid} synthesizes users with diverse intents to fine-tune LLMs accordingly. While these methods effectively identify intents, they are typically limited to inference-time prompting, without integrating intent into the model’s reasoning and training process, resulting in relatively surface-level intent modeling; they also often depend on multi-turn conversational context that is not always available~\citep{arora-etal-2024-intent, 10.1609/aaai.v39i23.34688}.

\section{Methodology}

This section presents IAP, our proposed framework for intent-aware PQA. Given a question, IAP infers the user’s latent intents, incorporates this intent into the model’s intermediate reasoning process, and generates an intent-aware personalized response. The framework is trained with RL under personalized reward feedback using a DAPO-based optimization procedure. Figure~\ref{fig:framework1} provides an overview of the framework. The following subsections describe the rollout sampling, personalized reward design, and policy optimization.

\subsection{Problem Formulation.}

The training dataset is defined as $D=\{(x_i, E_{x_i})\}_{i=1}^{|D|}$, where $x_i$ denotes the user question (including its narrative context), and $E_{x_i}=\{e_j\}_{j=1}^{|E_{x_i}|}$ is a set of personalized rubric aspects that a response to $x_i$ is expected to address. Given $x_i$, the objective is to generate a context-aware response $\hat{y}_i$ that aligns with the user's needs and goals. Following~\citet{salemi-zamani-2025-lamp}, we evaluate $\hat{y}_i$ using the rubric aspects $E_{x_i}$. Specifically, we compute a scoring function $\mu(x_i, \hat{y}_i, E_{x_i})$ to measure the extent to which $\hat{y}_i$ adequately covers these aspects. Details of the evaluation procedure are provided in Section~\ref{sec:reward}. We also define an intent-free response $\bar{y}_i \sim \pi_{\text{ref}}(\cdot \mid x_i)$, generated by a frozen reference model $\pi_{\text{ref}}$ without any intent-conditioning, which is used later as a contrastive signal during reward computation (Section~\ref{sec:reward}). IAP optimizes a policy $\pi_\theta$ via group-based sampling by the DAPO objective~\citep{yu2025dapo}. For each input $x_i$, the behavior policy samples a group of $G$ latent-intent rollouts $\{\hat{y}_i^{(g)}\}_{g=1}^{G} \sim \pi_{\theta_{\text{old}}}(\cdot \mid x_i).$ Each rollout $\hat{y}_i^{(g)}$ is a structured generation that contains an explicit latent-intent hypothesis and a corresponding personalized answer segment, denoted $\hat{y}_{i,g}$. We parameterize $\pi_\theta$ with an LLM initialized from a pretrained checkpoint and regularized by a frozen reference model $\pi_{\text{ref}}$.

\subsection{Latent-Intent Rollout Sampling Strategy}
\label{sec:sampling}

This section describes the rollout sampling procedure for IPA. Following group-based rollout sampling in DAPO training~\citep{rafailov2023direct, shao2024deepseekmath, yu2025dapo}, for each training instance $x_i$, where $x_i$ is the user question and its associated narrative, the current policy samples a group of $G$ intent-conditioned rollouts $\{\hat{y}_i^{(g)}\}_{g=1}^{G} \sim \pi_\theta(\cdot \mid x_i).$ Each rollout $\hat{y}_i^{(g)}$ infers plausible underlying intents, performs intermediate thinking and reasoning, and generates a personalized response conditioned on the inferred intents. Each rollout is generated using a prompt that constrains outputs to a tagged format. The model reasons about the user question and narrative inside an initial {\color{blue}\texttt{<think>}} and {\color{blue}\texttt{</think>}} block to hypothesize potential underlying intent(s). It then outputs a non-empty numbered list of candidate latent intents enclosed by {\color{red}\texttt{<intent>}} and {\color{red}\texttt{</intent>}}, where each intent captures a plausible hidden motivation or need. A second thinking phase follows inside another {\color{blue}\texttt{<think>}} and {\color{blue}\texttt{</think>}} block, explicitly conditioning on the inferred intents to plan the personalized response. The generation concludes with a personalized answer enclosed by {\color{teal}\texttt{<answer>}} and {\color{teal}\texttt{</answer>}}. The prompt is as follows:
\vspace{-0.5cm}
\medskip
\begin{adjustwidth}{1em}{1em}
\footnotesize
\setstretch{1}
\noindent
\hrule
Your task is to generate a personalized response to the user's question based on users' intent. To do this, you need to first think about the question and how to generate a personalized answer for the user based on the user intent and needs. The thinking process should be inside {\color{blue}\texttt{<think>}} and {\color{blue}\texttt{</think>}} tags. Then, generate the potential intent(s) as a numbered list enclosed within the {\color{red}\texttt{<intent>}} and {\color{red}\texttt{</intent>}} tags. Next, think again using those intent(s) to guide answer planning and generation, again inside {\color{blue}\texttt{<think>}} and {\color{blue}\texttt{</think>}} tags and put the final answers in the {\color{teal}\texttt{<answer>}} and {\color{teal}\texttt{</answer>}} tags. Nothing should be outside the mentioned tags except the initial question.
Now, answer the following question: \textbf{Question}.
\vspace{0.3cm}
\hrule
\end{adjustwidth}
\medskip

At the end of rollout sampling, we obtain a set of rollouts for the instance $x_i$, denoted $Y_i = \{\hat{y}_i^{(1)}, \ldots, \hat{y}_i^{(G)}\}$. 
%During training, we compute the personalized reward on the answer segment $\hat{y}_{i,g}$, namely the content enclosed by {\color{teal}\texttt{<answer>}} and {\color{teal}\texttt{</answer>}}, and use within-group comparisons across $\{\hat{y}_i^{(g)}\}_{g=1}^{G}$ to form advantages for policy optimization (Section~\ref{sec:adv}). The full structured rollout $\hat{y}_i^{(g)}$, including the inferred intent(s), is retained during optimization to couple latent-intent inference with downstream answer generation.

\subsection{Personalized Reward Modeling.}\label{sec:reward}

The reward function provides the main supervision signal for policy optimization in IAP. Following answer-level evaluation strategies in prior work~\citep{jin2025searchr, amirizaniani2026learning}, rewards are computed based on the quality of the final response in the {\color{teal}\texttt{<answer>}} tag, rather than intermediate outputs such as the inferred intents in the {\color{red}\texttt{<intent>}} tag. This design enables IAP to learn intent-aware generation without requiring gold intent labels. The overall reward is defined as a weighted combination of three components:
\begin{equation}
r_{i,g} = \lambda_1 r^{\text{pers}}_{i,g} - \lambda_2 r^{\text{contrast}}_{i,g} - \lambda_3 r^{\text{length}}_{i,g},
\end{equation}
where $\lambda_1$, $\lambda_2$, and $\lambda_3$ weight the three reward components, defined as follows:

\paragraph{Personalized Reward ($r^{\text{pers}}$).}
The primary reward signal is based on the LaMP-QA personalized evaluation metric~\citep{salemi-zamani-2025-lamp}, which evaluates a generated response against a set of user-specific rubric aspects $E_{x_i}$ that reflect how well the response addresses the user's underlying needs and goals. We instantiate the evaluator using Qwen2.5-32B-Instruct~\citep{qwen2.5} with temperature 0.0 as the judging model. For each query-response pair $(x_i, \hat{y}_{i,g})$, the evaluator assigns discrete aspect-level scores that are rescaled to $[0,1]$ and aggregated across all aspects. The personalized reward is therefore defined as: $r^{\text{pers}}_{i,g} = \mu(x_i, \hat{y}_{i,g}, E_{x_i}),$ where $\mu(\cdot)$ denotes the aggregated personalized evaluation score.

\paragraph{Contrastive Intent-Free Reward ($r^{\text{contrast}}$).}
To make the reward more informative, we introduce a contrastive intent-free reward following~\citep{10.1145/3746252.3760851, amirizaniani2026learning}. Specifically, we generate an intent-free alternative for each query and penalize intent-aware responses that fail to distinguish themselves from it, either by remaining semantically similar or by scoring no higher on personalization quality. By doing so, it discourages generic generation and guides the model toward responses that are more sensitive to the user's inferred intent and therefore more personalized. For each input question $x_i$, the intent-free answer $\bar{y}_i \sim \pi_{\text{ref}}(\cdot \mid x_i)$ is generated by the frozen reference policy $\pi_{\text{ref}}$. The contrastive reward is then defined as: $r^{\text{contrast}}_{i,g} = \lambda \cdot \text{BERTScore}(\hat{y}_{i,g}, \bar{y}_i) + \max\left(0, \mu(\bar{y}_i) - \mu(\hat{y}_{i,g})\right),$ where $\text{BERTScore}(\hat{y}_{i,g}, \bar{y}_i)$ measures the semantic similarity between the generated answer and the intent-free reference using BERTScore(\texttt{microsoft/deberta-v3-large})~\citep{zhang2020bertscore}, and $\max\left(0, \mu(\bar{y}_i) - \mu(\hat{y}_{i,g})\right)$ penalizes the model when its answer fails to outperform the intent-free response. Together, these two terms are minimized when the generated response is semantically distinct from and of higher personalization quality than the intent-free answer, and maximized otherwise, thereby discouraging generic, non-intent-aware generation.

\paragraph{Intent Length Regularization ($r^{\text{length}}$).}
To discourage verbose intent generation and promote concise intent representations, we apply a length-based regularization term $r^{\text{length}}$ to the \texttt{<intent>} tag, inspired by~\citep{yu2025dapo, guo2025deepseek}. Concretely, when the generated intent exceeds a predefined threshold $L$, the excess length is penalized proportionally: $r^{\text{length}}_{i,g} = \max(0, |\texttt{<intent>}| - L).$ This design encourages the model to express the user’s intents in a concise form, rather than through unnecessarily verbose intent descriptions.
\begin{algorithm}[t]
\caption{IAP: Intent-Aware Personalization for Question Answering.}
\label{alg:ic-dapo}
\footnotesize
\textbf{Input:} policy model $\pi_\theta$; frozen reference model $\pi_{\text{ref}}$; 
LaMP-QA evaluator $\mu$; training set $\mathcal{D}=\{(x_i,E_{x_i})\}$; 
hyperparameters $G, \lambda_1, \lambda_2, \lambda_3, \lambda, L, \mu_{\text{upd}}$ \\
\For{step $=1,\ldots,M$}{
    Sample a batch of training instances $\mathcal{D}_b$ from $\mathcal{D}$\;
    Update the behavior policy $\pi_{\theta_{\text{old}}} \leftarrow \pi_\theta$\;
    \ForEach{$(x_i, E_{x_i}) \in \mathcal{D}_b$}{
        Generate intent-free answer $\bar{y}_i \sim \pi_{\text{ref}}(\cdot \mid x_i)$\;
        Sample $G$ rollouts $\{\hat{y}_{i,g}\}_{g=1}^{G} \sim \pi_{\theta_{\text{old}}}(\cdot \mid x_i)$\;
        \For{$g=1$ \KwTo $G$}{
            Compute personalized reward $r^{\text{pers}}_{i,g}=\mu(x_i,\hat{y}_{i,g},E_{x_i})$\;
            Compute contrastive intent-free reward $r^{\text{contrast}}_{i,g} = \lambda \cdot \text{BERTScore}(\hat{y}_{i,g}, \bar{y}_i) + \max\left(0, \mu(\bar{y}_i) - \mu(\hat{y}_{i,g})\right)$\;
            Compute intent length regularization $r^{\text{length}}_{i,g}=\max(0,|\texttt{<intent>}_{i,g}|-L)$\;
            Compute final reward $r_{i,g}=\lambda_1 r^{\text{pers}}_{i,g} - \lambda_2 r^{\text{contrast}}_{i,g} - \lambda_3 r^{\text{length}}_{i,g}$\;
        }
        Compute group-relative advantages from $\{r_{i,g}\}_{g=1}^{G}$\;
        \For{update $=1,\ldots,\mu_{\text{upd}}$}{
            Update $\pi_\theta$ using the DAPO objective over the sampled rollouts\;
        }
    }
}
\textbf{Output:} Optimized policy $\pi_\theta$
\end{algorithm}
\subsection{Group-Relative Advantage and DAPO Objective Updates.}\label{sec:adv}
For group computation and policy optimization, IAP follows the original DAPO formulation in~\citep{yu2025dapo}. Specifically, we use the same group-relative advantage computation and policy update objective as defined in the original work, without modifying the underlying optimization equations. Using the reward defined in Section~\ref{sec:reward}, DAPO is applied directly to optimize sampled rollouts. Thus, our adaptation lies not in the optimization algorithm itself, but in the design of the reward signal. Algorithm~\ref{alg:ic-dapo} summarizes IAP.

\section{Experiment}

\paragraph{\textbf{Datasets and Evaluation.}}
We conduct our experiments on the LaMP-QA~\citep{salemi-zamani-2025-lamp}, a dataset for PQA spanning three domains: Art \& Entertainment, Lifestyle \& Personal Development, and Society \& Culture~\footnote{For brevity, in this paper, we refer to \textit{Art \& Entertainment} as ``Art,'' \textit{Lifestyle \& Personal Development} as ``Lifestyle,'' and \textit{Society \& Culture} as ``Society.''}. Each instance includes a user query, a narrative of the user’s intent and preferences, and personalized rubrics describing what an ideal response should address. Following~\citet{salemi-zamani-2025-lamp}, we evaluate responses by how well they satisfy user-specific rubric aspects, using Qwen2.5-32B-Instruct~\citep{qwen2.5} at temperature 0.0. Each response is scored on personalized rubric aspects from 0 to 2, normalized to $[0,1]$, and averaged across aspects. This yields a scalar score measuring personalized answer quality through alignment with user-specific criteria.

\paragraph{\textbf{Baselines.}}
To evaluate IAP, we consider several baselines: (1) \textbf{Non-personalized}, a CoT-based No-Personalization setting following~\citep{salemi-zamani-2025-lamp}, where the model generates responses without any user information; (2) \textbf{Implicit Intent Personalization}, where the model implicitly infers intent through CoT prompting before generating a personalized response, without any explicit intent signal\footnotemark; 
(3) \textbf{Explicit Intent Personalization}, where the intent predicted by IAP, represented by the \texttt{<intent>} tag, is explicitly included in the CoT-based prompt to guide response generation\footnotemark[\value{footnote}];and (4) \textbf{Supervised Fine-Tuning (SFT)}~\citep{wei2022finetuned}, where the model generates five answers per question, selects the one with the highest personalized score, and uses these silver-labeled pairs for fine-tuning.
%Together, these baselines isolate the effects of personalization, inference-time intent conditioning, and full intent-aware optimization.

\footnotetext{Prompts are provided in Appendix~\ref{prompt}.}

\paragraph{\textbf{IAP Training Configuration.}}
We evaluate six variants from Qwen2.5-7B~\citep{qwen2.5}, Gemma3-12B~\citep{gemmateam2025gemma3}, and Mistral-Small-24B~\footnote{\url{https://huggingface.co/mistralai/Mistral-Small-24B-Instruct-2501}}, including base and instruction-tuned models, all trained on the LaMP-QA training split~\citep{salemi-zamani-2025-lamp}. Following prior work~\citep{10.1145/3746252.3760851, jin2025searchr}, we train each model with 5 rollouts per instance (Details in the Appendix~\ref{group}). Training runs for 200 steps with a batch size of 512 using TRL~\footnote{\url{https://hf.co/docs/trl/en/grpo_trainer}}
, a learning rate of $1\times10^{-5}$, and a warm-up ratio of 0.285. To improve memory efficiency, we enable gradient checkpointing and use FSDP. For rollout generation, we use vLLM~\footnote{\url{https://github.com/vllm-project/vllm}} with tensor parallel size 1, GPU memory utilization 0.6, temperature 1.0, and top-$p=1.0$. We fix the KL regularization coefficient $\beta=0.001$ and clipping parameter $\epsilon=0.2$ across all experiments. All IAP and baseline runs use NVIDIA H100 GPUs (80GB VRAM, 512GB RAM), a 2048-token budget, and a 400-token limit for \texttt{<intent>}.
%We save checkpoints every 100 steps and use the final checkpoint for evaluation unless training diverges, in which case we select the most recent stable checkpoint according to the training reward curve.

\section{Results and Analysis}

\subsection{Comparison with the Baselines}
Table~\ref{tab:main-results} compares IAP with baselines on the LaMP-QA test set, where it achieves the best performance, demonstrating its effectiveness in generating intent-aware personalized responses. While intent-based prompting improves over no personalization, its gains are limited, and SFT, though often the strongest baseline, lags behind IAP. To better quantify IAP’s advantage, Table~\ref{tab:main-results} reports delta values (shown in parentheses in the IAP row) that indicate its improvement over the strongest baseline (underlined). Positive deltas across nearly all LLMs and benchmarks show that IAP consistently improves intent-aware personalized answer generation.
%Average relative improvements across the three domains and macro score are 9.95\% for Qwen2.5-7B-it, 8.44\% for Qwen2.5-7B, 6.07\% for Gemma3-12B-it, 4.47\% for Gemma3-12B, 10.22\% for Mistral-Small-24B-it, and 6.80\% for Mistral-Small-24B. 
\begin{table*}[t]
\centering
\scriptsize
\setlength{\tabcolsep}{6pt}
\renewcommand{\arraystretch}{1.15}
\caption{Performance comparison across models and baselines. Best results are in \textbf{bold}, second-best are \underline{underlined}, and * marks significance over all baselines ($p<0.01$).}
\vspace{-0.2cm}
\begin{tabular}{l|l|lll|l}
\hline
\textbf{Model} & \textbf{Method} & \textbf{Art} & \textbf{Lifestyle} & \textbf{Society} & \textbf{Avg. (macro)} \\
\hline

\multirow{6}{*}{Qwen2.5-7B-it}
& Non-pers.                  & 0.2929 & 0.4083 & 0.4469 & 0.3827 \\
& Implicit Intent Pers. & 0.3004 & 0.4206 & 0.4729 & 0.3979 \\
& Explicit Intent Pers.   & 0.3032 & 0.4266 & 0.4766 & 0.4021 \\
& SFT                                & \underline{0.3158} & \underline{0.4311} & \underline{0.4801} & \underline{0.4090} \\
\cdashline{2-6}
& IAP(Ours)                               & \textbf{0.3533*} (+11.87\%) & \textbf{0.4718*} (+9.44\%) & \textbf{0.5240*} (+9.14\%) & \textbf{0.4497*} (+9.95\%) \\
\hline

\multirow{6}{*}{Qwen2.5-7B}
& Non-pers.                  & 0.3129 & 0.4582 & 0.4769 & 0.4160 \\
& Implicit Intent Pers. & 0.3158 & 0.4639 & 0.4777 & 0.4191 \\
& Explicit Intent Pers.   & 0.3169 & \underline{0.4658} & 0.4803 & \underline{0.4210} \\
& SFT                                & \underline{0.3221} & 0.4526 & \underline{0.4837} & 0.4194 \\
\cdashline{2-6}
& IAP(Ours)                               & \textbf{0.3629*} (+12.67\% ) & \textbf{0.4889*} (+4.96\%) & \textbf{0.5178*} (+7.05\%) & \textbf{0.4565*} (+8.44\%) \\
\hline

\multirow{6}{*}{Gemma3-12B-it}
& Non-pers.                  & 0.4130 & \underline{0.5952} & \underline{0.6292} & \underline{0.5458} \\
& Implicit Intent Pers. & 0.4211 & 0.5872 & 0.6044 & 0.5375 \\
& Explicit Intent Pers.   & 0.4270 & \textbf{0.5982} & 0.6088 & 0.5446 \\
& SFT                                & \underline{0.4289} & 0.5801 & 0.6118 & 0.5402 \\
\cdashline{2-6}
& IAP(Ours)                               & \textbf{0.4769*} (+11.19\%) & 0.5934 (-0.8\%) & \textbf{0.6665*} (+5.93\%) & \textbf{0.5789*} (+6.07\%) \\
\hline

\multirow{6}{*}{Gemma3-12B}
& Non-pers.                 & 0.4033 & 0.6005 & 0.6048 & 0.5362 \\
& Implicit Intent Pers. & 0.4078 & 0.6034 & 0.6120 & 0.5410 \\
& Explicit Intent Pers.  & 0.4155 & 0.6129 & 0.6138 & 0.5474 \\
& SFT                                & \underline{0.4230} & \underline{0.6144} & \underline{0.6147} & \underline{0.5507} \\
\cdashline{2-6}
& IAP(Ours)                              & \textbf{0.4612*} (+9.03\%) & \textbf{0.6218} (+1.20\%) & \textbf{0.6430*} (+4.60\%) & \textbf{0.5753*} (+4.47\%) \\
\hline

\multirow{6}{*}{\shortstack{Mistral-Small\\24B-it}}
& Non-pers.                  & 0.3966 & 0.5610 & 0.6003 & 0.5193 \\
& Implicit Intent Pers. & 0.4028 & 0.5738 & 0.6049 & 0.5271 \\
& Explicit Intent Pers.   & 0.4047 & 0.5788 & 0.6072 & 0.5302 \\
& SFT                                & \underline{0.4161} & \underline{0.5878} & \underline{0.6157} & \underline{0.5398} \\
\cdashline{2-6}
& IAP(Ours)                               & \textbf{0.4811*} (+15.62\%) & \textbf{0.6505*} (+10.67\%) & \textbf{0.6536*} (+6.16\%) & \textbf{0.5950*} (+10.22\%) \\
\hline

\multirow{6}{*}{\shortstack{Mistral-Small\\24B}}
& Non-pers.                  & 0.4027 & 0.5812 & 0.5926 & 0.5255 \\
& Implicit Intent Pers. & 0.4158 & 0.6033 & 0.6184 & 0.5458 \\
& Explicit Intent Pers.   & 0.4189 & 0.6212 & 0.6255 & 0.5552 \\
& SFT                                & \underline{0.4359} & \underline{0.6247} & \underline{0.6430} & \underline{0.5678} \\
\cdashline{2-6}
& IAP(Ours)                               & \textbf{0.4918*} (+12.82\%) & \textbf{0.6649*} (+6.44\% ) & \textbf{0.6628} (+3.08\%) & \textbf{0.6065*} (+6.80\%) \\
\hline
\end{tabular}
\label{tab:main-results}
\end{table*}
The only exception is a slight drop of 0.8\% for Gemma3-12B-it on Lifestyle. This result is further supported by statistical testing. As shown by * in Table~\ref{tab:main-results}, almost all improvements remain significant under a Bonferroni-corrected threshold of $p<0.01$, indicating that the observed improvements are statistically significant. Together, these results show that IAP improves intent-aware PQA performance over all baselines.

\subsection{Training Dynamics of IAP}
\begin{figure}[t]
    \centering
    \begin{subfigure}[t]{0.32\textwidth}
        \centering
        \includegraphics[width=\linewidth]{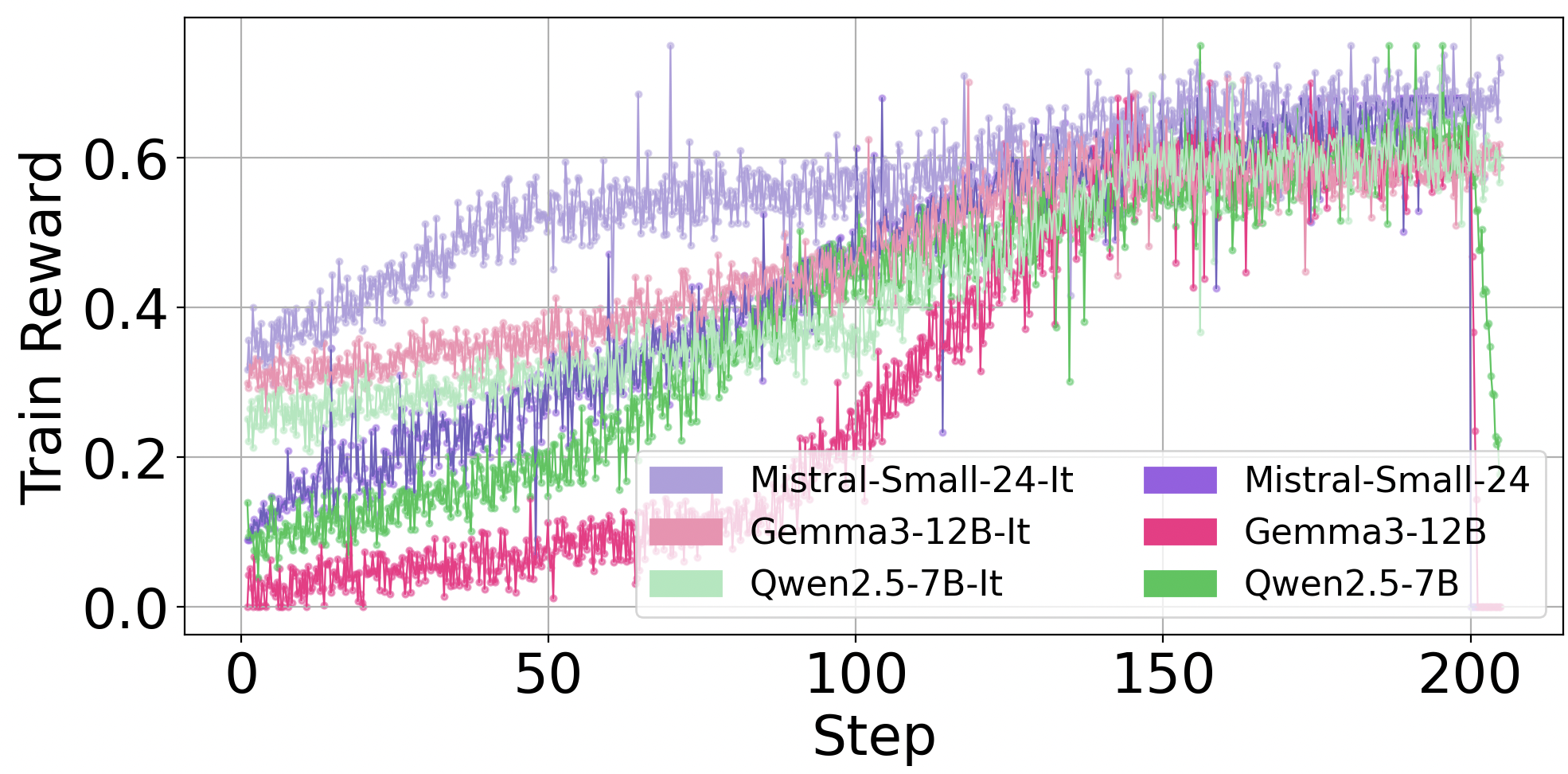}
        \vspace{-0.3cm}
        \caption{Training process.}
        \label{Fig:Training process}
    \end{subfigure}
    \hfill
    \begin{subfigure}[t]{0.32\textwidth}
        \centering
        \includegraphics[width=\linewidth]{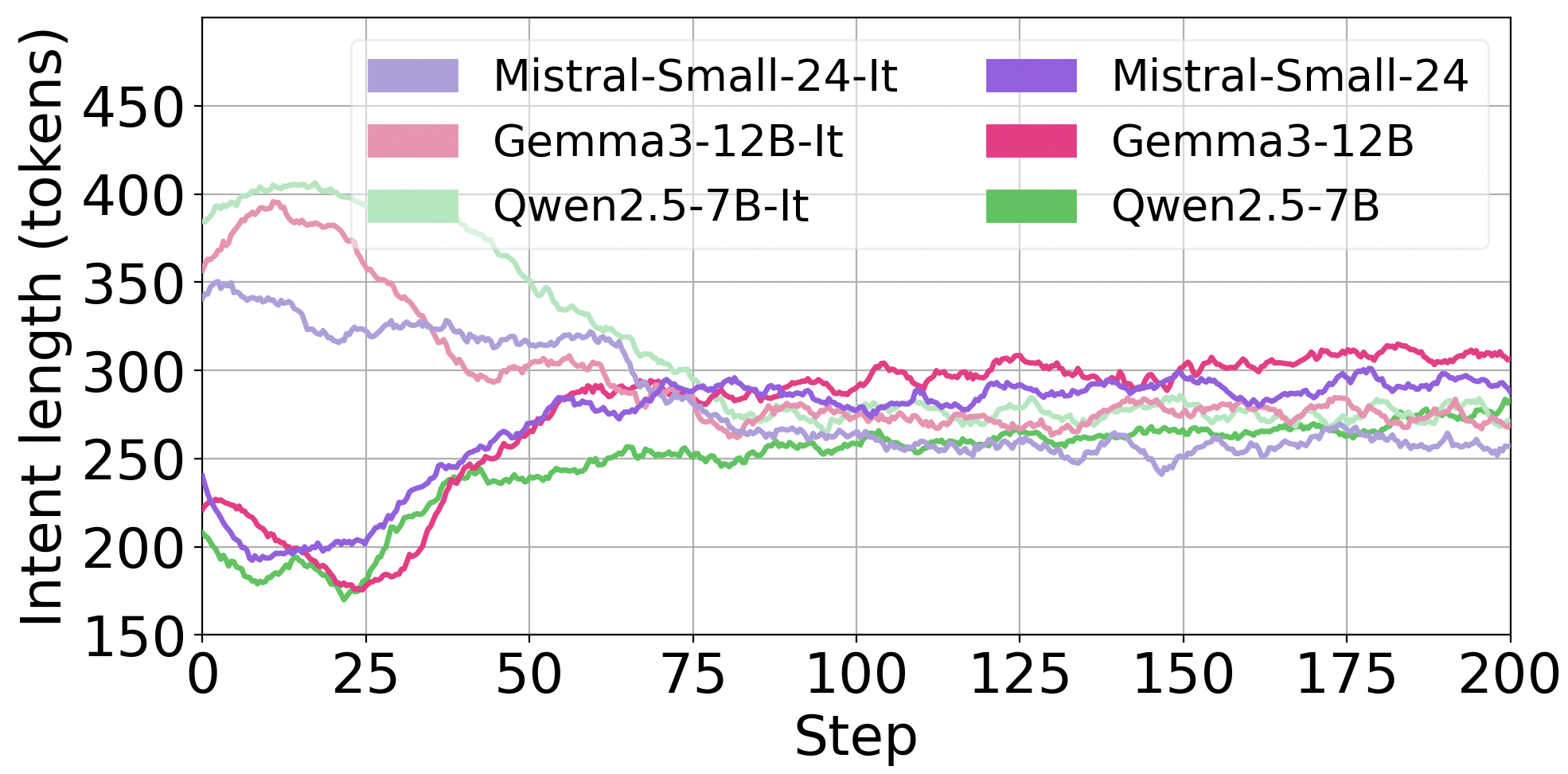}
        \vspace{-0.3cm}
        \caption{Intent length.}
        \label{Fig:int-len}
    \end{subfigure}
    \hfill
    \begin{subfigure}[t]{0.32\textwidth}
        \centering
        \includegraphics[width=\linewidth]{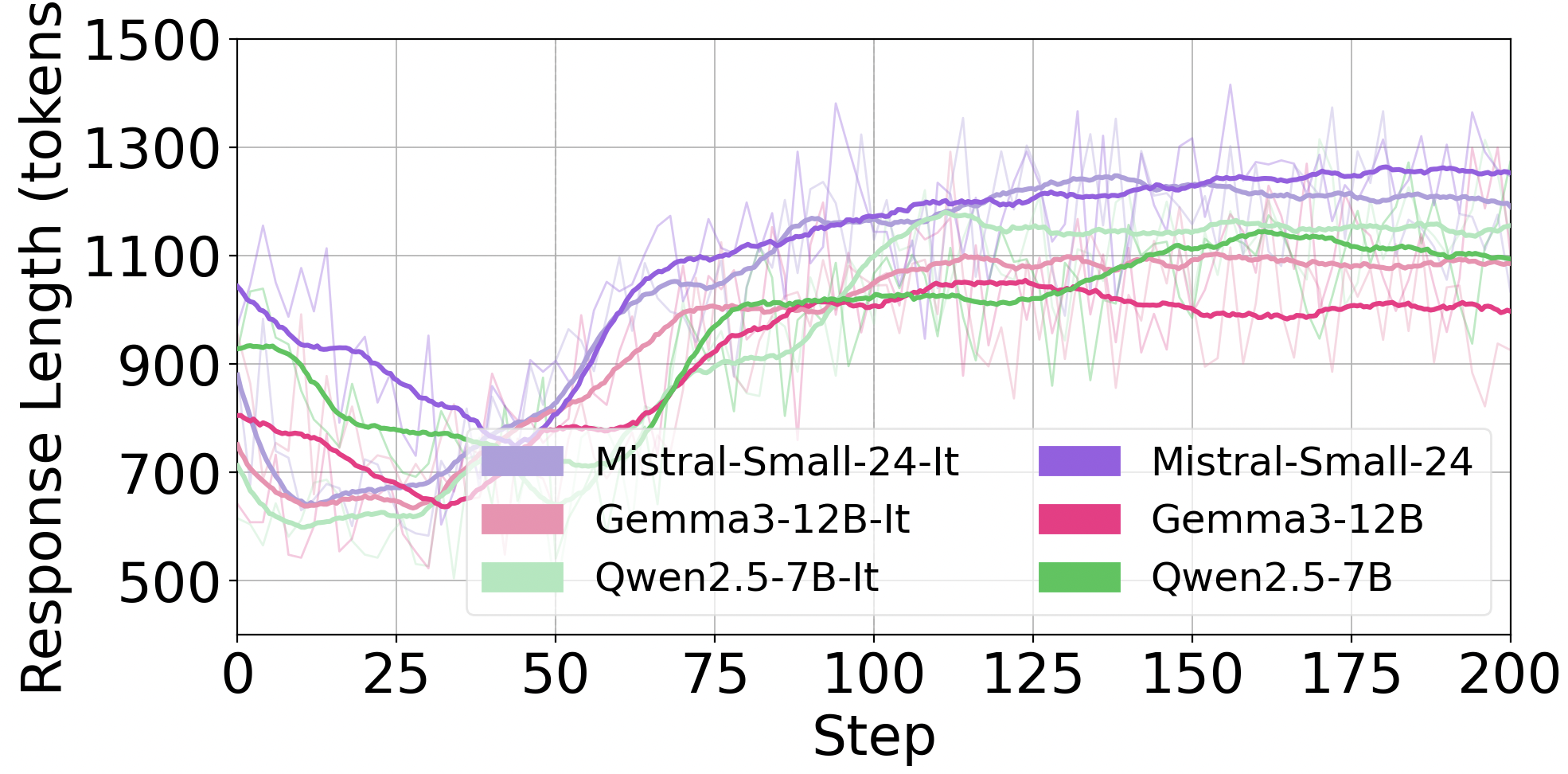}
        \vspace{-0.3cm}
        \caption{Response length.}
        \label{Fig:res-len}
    \end{subfigure}
    \vspace{-0.2cm}
    \caption{The log of reward, intent and respond length of IAP on the LAMP-QA.}
    \label{fig:three_plots}
\end{figure}

\paragraph{\textbf{Training Reward.}}
Figure~\ref{Fig:Training process} shows reward trajectories over training steps for all LLMs. Rewards rise quickly early in training, indicating rapid learning of intent-aware personalized responses, and then stabilize at higher levels with smaller gains over time. This pattern suggests that the largest gains occur early in training, when the models are learning to better capture the intent behind the question, followed by a more stable phase in which this ability is refined to produce stronger personalized responses. Although the curves remain noisy, with local fluctuations and occasional short-term drops, training stays stable overall. Notably, none of the models show reward collapse or sustained performance decline.
%Overall, these results indicate that the training procedure provides a stable and effective learning signal for improving intent-aware personalized response generation.
\paragraph{\textbf{Intent and Response Length.}}
Figures~\ref{Fig:int-len} and ~\ref{Fig:res-len} show that, during training, both intent and response lengths follow structured dynamics across LLMs. Intent lengths fluctuate in the early stages but gradually stabilize after approximately step 75, indicating that the models converge toward a more consistent way of representing inferred user intent. Response lengths exhibit a dip-then-rise pattern before reaching a more stable range, with some models stabilizing slightly later than others. Taken together, these trends suggest that training leads the models toward a more stable and controlled generation regime. 

\subsection{Effect of number of intent in IAP.}
\begin{wrapfigure}{r}{0.35\textwidth}

  \centering
  \includegraphics[width=0.40\textwidth]{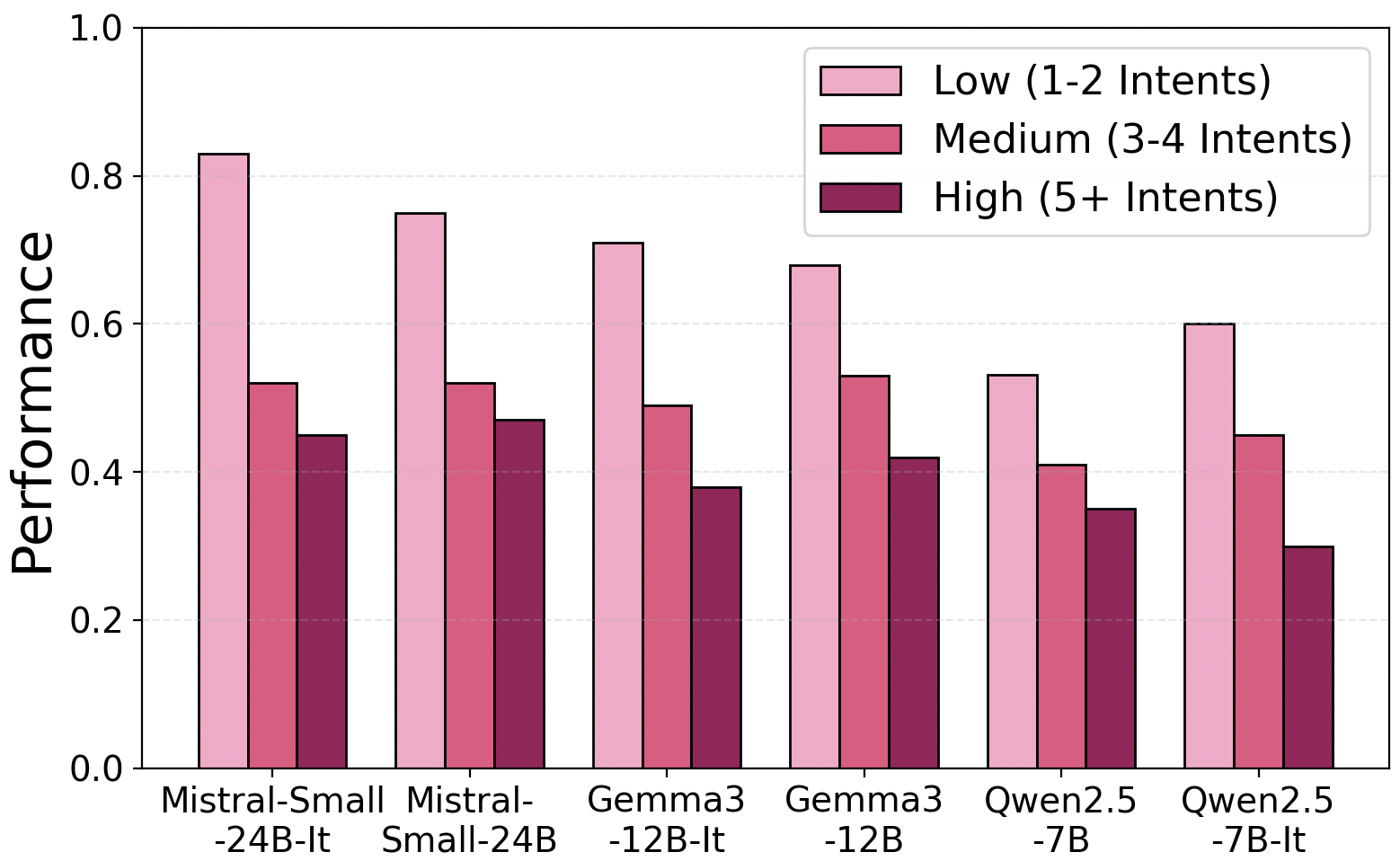}

  \caption{Number of intents.}
  \label{fig:intent}

\end{wrapfigure}

To better understand the role of the \texttt{<intent>} tag, we analyze performance on the LaMP-QA test set as a function of the number of inferred intents per query, grouped into Low (1--2), Medium (3--4), and High (5+). Figure~\ref{fig:intent} shows that average performance is highest in the low-intent bucket and lower in the medium- and high-intent buckets. Although performance varies across LLMs, the same ordering holds throughout, indicating that answer quality declines as the number of inferred intents increases. This pattern suggests that response generation becomes more challenging as the number of inferred intents increases.

\section{Ablation Study}
%This section ablates IAP’s two core components: the \texttt{<intent>} tag and the contrastive intent-free reward.
\begin{table*}[t]
\centering
\scriptsize
\setlength{\tabcolsep}{5pt}
\renewcommand{\arraystretch}{1.15}
\caption{Performance under different intent settings.}
\vspace{-0.3cm}
\begin{tabular}{l|l|cc|cc|cc}
\hline
\textbf{Model} & \textbf{Intent Setting} 
& \multicolumn{2}{c|}{\textbf{Art}}
& \multicolumn{2}{c|}{\textbf{Lifestyle}}
& \multicolumn{2}{c}{\textbf{Society}} \\
\cline{3-8}
& 
& \textbf{Reward} & \textbf{$\Delta$}
& \textbf{Reward} & \textbf{$\Delta$ }
& \textbf{Reward} & \textbf{$\Delta$} \\
\hline

\multirow{4}{*}{\shortstack{Qwen2.5-\\7B-it}}
& No Intent                 & 0.0653 & -0.3445 & 0.0956 & -0.4121 & 0.0733 & -0.4811 \\
& In-group     & 0.2028 & -0.2070 & 0.1923 & -0.3154 & 0.2618 & -0.2926 \\
& Cross-group  & 0.0872 & -0.3226 & 0.1549 & -0.3528 & 0.1439 & -0.4105 \\
& Selected-intent           & \textbf{0.4098} & -       & \textbf{0.5077} & -      & \textbf{0.5544} & -       \\
\hline

\multirow{4}{*}{Qwen2.5-7B}
& No Intent                 & 0.0611 & -0.3672 & 0.0904 & -0.4134 & 0.0848 & -0.4117 \\
& In-group     & 0.1169 & -0.3114 & 0.2213 & -0.2825 & 0.2022 & -0.2943 \\
& Cross-group  & 0.0975 & -0.3308 & 0.0821 & -0.4217 & 0.1069 & -0.3896 \\
& Selected-intent           & \textbf{0.4283} & -       & \textbf{0.5038} & -       & 0.4965 & -       \\
\hline

\multirow{4}{*}{\shortstack{Gemma3-\\12B-it}}
& No Intent                 & 0.0822 & -0.4252 & 0.1084 & -0.5437 & 0.1168 & -0.5591 \\
& In-group   & 0.2239 & -0.2835 & 0.2819 & -0.3702 & 0.3504 & -0.3255 \\
& Cross-group  & 0.0751 & -0.4323 & 0.1120 & -0.5401 & 0.1033 & -0.5726 \\
& Selected-intent           & \textbf{0.5074} & -       & \textbf{0.6521} & -       & \textbf{0.6759} & -       \\
\hline

\multirow{4}{*}{Gemma3-12B}
& No Intent                 & 0.1160 & -0.3989 & 0.0850 & -0.5369 & 0.0764 & -0.6066 \\
& In-group    & 0.2047 & -0.3102 & 0.1650 & -0.4569 & 0.1437 & -0.5393 \\
& Cross-group  & 0.1547 & -0.3602 & 0.1436 & -0.4783 & 0.1219 & -0.5611 \\
& Selected-intent           & \textbf{0.5149} & -       & \textbf{0.6219} & -       & \textbf{0.6830} & -       \\
\hline

\multirow{4}{*}{\shortstack{Mistral-Small\\24B-it}}
& No Intent                 & 0.1243 & -0.4112 & 0.1688 & -0.5252 & 0.1529 & -0.5595 \\
& In-group     & 0.1669 & -0.3686 & 0.2014 & -0.4926 & 0.2068 & -0.5056 \\
& Cross-group  & 0.1099 & -0.4256 & 0.1875 & -0.5065 & 0.1320 & -0.5804 \\
& Selected-intent           & \textbf{0.5355} & -       & \textbf{0.6940} & -       & \textbf{0.7124} & -       \\
\hline

\multirow{4}{*}{\shortstack{Mistral-Small\\24B}}
& No Intent                 & 0.0976 & -0.4242 & 0.1159 & -0.5591 & 0.0872 & -0.5857 \\
& In-group     & 0.1549 & -0.3669 & 0.1811 & -0.4939 & 0.1827 & -0.4902 \\
& Cross-group  & 0.1370 & -0.3848 & 0.1658 & -0.5092 & 0.1529 & -0.5200 \\
& Selected-intent           & \textbf{0.5218} & -       & \textbf{0.6750} & -       & \textbf{0.6729} & -       \\
\hline
\end{tabular}
\label{tab:intent_ablation_rewards}
\end{table*}
\paragraph{\textbf{ Ablation of the \texttt{<intent>} tag in IAP.}}
%To evaluate the contribution of the inferred \texttt{<intent>}, we compare the selected-intent setting with three alternative rollout conditions (Table~\ref{tab:intent_ablation_rewards}). For each condition, we generate IAP rollout trajectories using the prompts in Appendix~\ref{prompt} and evaluate the outputs with the personalized reward model in Section~\ref{sec:reward}. The alternatives are: \textit{No Intent}, where no intent signal is provided; \textit{In-group random intent}, where the model is conditioned on a randomly selected intent from the same LaMP-QA category and domain; and \textit{Cross-group random intent}, where the intent is drawn from a different category and domain. 
To evaluate the contribution of the inferred \texttt{<intent>}, we take the fully trained IAP model and replace the inferred intent at inference time with three alternative (Table~\ref{tab:intent_ablation_rewards}). Each alternative is fed to the model using the prompts in Appendix~\ref{prompt}, with responses evaluated using the personalized reward model in Section~\ref{sec:reward}. By fixing the policy weights and varying only the intent signal, this ablation directly isolates the effect of intent quality on response generation. \textit{Selected-intent} is the intent inferred by IAP, and the three ablations are: \textit{No Intent}, where no intent signal is provided; \textit{In-group random intent}, where the model is conditioned on a random intent from the same LaMP-QA category; and \textit{Cross-group random intent}, where the intent is drawn from a different category. The selected-intent setting yields the highest personalized score across all LLMs (Table~\ref{tab:intent_ablation_rewards}). These large gaps suggest that the gains come from conditioning generation on the correct intent. The delta values further clarify this effect. Each delta is computed as the difference between the average reward score of a given setting and that of the selected-intent condition, measuring the performance drop when the correct intent is missing or replaced. Larger negative deltas indicate greater reliance on the correct \texttt{<intent>} signal for personalized response quality. As shown in Table~\ref{tab:intent_ablation_rewards} removing intent leads to large drops, and random intents also perform far below the selected-intent setting. Overall, these results indicate that the inferred \texttt{<intent>} captures the user’s underlying intent.

%As shown in Table~\ref{tab:intent_ablation_rewards}, removing intent entirely leads to substantial drops, such as $-0.3445,  -0.4121,  -0.4811$ for Qwen2.5-7B-it and $-0.4112, -0.5252, -0.5595$ for Mistral-Small-24B-it across Art, Lifestyle, and Society. Even when the model is given a random intent, performance remains clearly below the selected-intent setting. For instance, In-group random intent still trails the selected intent by $-0.2070$ to $-0.3154$ on Qwen2.5-7B-it, while Cross-group random intent often causes even larger drops. Overall, these results show that <intent> serves as an effective intermediate representation for capturing the user’s underlying goal, and that correctly inferred intent is a key factor behind the stronger personalization performance of IAP.

\paragraph{\textbf{ Ablation on Contrast Intent-free Reward.}}
Table~\ref{tab:reward_wo_vanilla_vs_icqa} presents an ablation study on the intent-free contrastive reward, showing that the full IAP framework consistently outperforms the variant without this component (\textit{contrast intent-free reward}). In IAP, the reward favors intent-aware responses by subtracting the intent-free signal during optimization. As a result, the model is encouraged to prefer answers that are more personalized, rather than generic responses. In other words, without the contrastive reward, the model has no reason to prefer a personalized answer over a generic one that scores equally well. Introducing the reward created an intent-free baseline, or a lower bound, that the reward must exceed.

\section{Discussion and Answers to Research Questions}
\paragraph{\textbf{Learning Latent Intent for Personalized Answer Generation (RQ1).}}
IAP is a tag-based framework for PQA that infers latent user intents, incorporates them into the model’s thinking and reasoning process, and trains LLMs to generate intent-aware personalized responses. IAP consistently outperforms all baselines (Table~\ref{tab:main-results}), suggesting that intent information is most effective when integrated into the model’s thinking process rather than provided only as an external cue, findings similar to~\citep{zhao2026improving}. The ablation study in Table~\ref{tab:intent_ablation_rewards} further reinforces this finding: replacing the correctly inferred \texttt{<intent>} tag with either no intent or a randomly selected one leads to substantial performance drops across all models, indicating that personalized response generation depends strongly on the quality and specificity of the inferred intent signal. This is consistent with prior work showing that understanding query-specific intent is key to improving performance~\citep{zhao2026improving, kim2025ipqa, askari-etal-2025-solid}.
%Taken together, these findings show that IAP is an effective and scalable approach for intent-aware PQA.
\begin{table*}[t]
\centering
\scriptsize
\setlength{\tabcolsep}{6pt}
\renewcommand{\arraystretch}{1.15}
\caption{Comparison between Rewarding without contrast intent-free and IAP.}
\vspace{-0.3cm}
\begin{tabular}{l|l|ccc|c}
\hline
\textbf{Model} & \textbf{Method} & \textbf{Art } & \textbf{Lifestyle} & \textbf{Society} & \textbf{Avg. (macro)} \\
\hline

\multirow{2}{*}{Qwen2.5-7B-it}
& Rewarding w.o. Contrast intent-free & 0.3207 & 0.4363 & 0.4971 & 0.4180 \\
& Rewarding w. Contrast intent-free                 & \textbf{0.3533} & \textbf{0.4718} & \textbf{0.5240} & \textbf{0.4497} \\
\hline

\multirow{2}{*}{Qwen2.5-7B}
& Rewarding w.o. Contrast intent-free & 0.3321 & 0.4426 & 0.4627 & 0.4124 \\
& Rewarding w. Contrast intent-free                  & \textbf{0.3629} & \textbf{0.4889} & \textbf{0.5178} & \textbf{0.4565} \\
\hline

\multirow{2}{*}{Gemma3-12B-it}
& Rewarding w.o. Contrast intent-free & 0.4207 & 0.5857 & 0.6210 & 0.5424 \\
& Rewarding w. Contrast intent-free                 & \textbf{0.4769} & \textbf{0.5934} & \textbf{0.6665} & \textbf{0.5789} \\
\hline

\multirow{2}{*}{Gemma3-12B}
& Rewarding w.o. Contrast intent-free & 0.4330 & 0.6117 & 0.6220 & 0.5555 \\
& Rewarding w. Contrast intent-free                  & \textbf{0.4912} & \textbf{0.6218} & \textbf{0.6430} & \textbf{0.5853} \\
\hline

\multirow{2}{*}{Mistral-Small-24B-it}
& Rewarding w.o. Contrast intent-free & 0.4261 & 0.5618 & 0.6087 & 0.5322 \\
& Rewarding w. Contrast intent-free                  & \textbf{0.4811} & \textbf{0.6505} & \textbf{0.6536} & \textbf{0.5950} \\
\hline

\multirow{2}{*}{Mistral-Small-24B}
& Rewarding w.o. Contrast intent-free & 0.4436 & 0.6222 & 0.6378 & 0.5678 \\
& Rewarding w. Contrast intent-free                  & \textbf{0.4918} & \textbf{0.6649} & \textbf{0.6628} & \textbf{0.6065} \\
\hline

\end{tabular}
\label{tab:reward_wo_vanilla_vs_icqa}
\vspace{-0.3cm}
\end{table*}
\paragraph{\textbf{Intent-Free Response Signal in Training (RQ2).}}
IAP’s reward design also plays a key role in intent-grounded personalized answer generation. The ablation study in Table~\ref{tab:reward_wo_vanilla_vs_icqa} provides clear evidence that explicitly accounting for intent-free responses during reward optimization is essential, as removing this component consistently degrades performance across all evaluated models. 
%While a model trained without this signal can still produce responses that partially satisfy personalized rubric criteria, it lacks an explicit incentive to move away from generic, non-intent-aware outputs.
Considering the intent-free response during optimization encourages the model to prefer answers that go beyond intent-free responses, leading to better alignment with the user’s underlying intent and contextual needs. This design is consistent with contrastive reward formulations in prior work, where contrastive outputs have been shown to promote more targeted and discriminative generation~\citep{10.1145/3746252.3760851, amirizaniani2026learning}. Broadly, these results show that IAP’s gains also stem from a reward design that discourages generic responses and favors intent-grounded ones.

\section{Conclusion}
This paper presents IAP, an RL framework that trains LLMs to infer implicit user intents from input questions and incorporate it into their thinking process to generate personalized, intent-grounded answers. Across six LLMs on LaMP-QA, IAP consistently outperforms all baselines, with an average macro-score gain of about 7.5\%. Taken together, these findings demonstrate the effectiveness of IAP in explicitly modeling the user’s implicit “why” and generating responses for single-turn PQA that better align with the user’s underlying goals.

%Unlike prior approaches that rely on fixed intent taxonomies, prompt-based inference, or rich multi-turn context, IAP is particularly suited for single-turn QA settings where intent must be inferred from limited textual input alone, without requiring any additional context beyond the input question. 

\bibliography{colm2026_conference}
\bibliographystyle{colm2026_conference}

\appendix
\section{Appendix}
\subsection{Prompt Instructions}\label{prompt}
Prompts used in the baseline settings for implicit and explicit intent personalization.
\paragraph{Implicit Intent Personalization}
\medskip
\begin{adjustwidth}{1.2em}{1.2em}
\footnotesize
\setstretch{0.95}
\noindent
\hrule
Your task is to generate a personalized response to the user’s question. First, infer the user’s intent, defined as the underlying goal, need, or reason behind asking the question. Then, use this understanding to produce a personalized response that is tailored to the user and directly addresses their needs. Think step by step. Now, answer the following question: \textbf{Question}.
\vspace{0.3cm}
\hrule
\end{adjustwidth}
\medskip
\paragraph{Explicit Intent Personalization}
\medskip
\begin{adjustwidth}{1.2em}{1.2em}
\footnotesize
\setstretch{0.95}
\noindent
\hrule
Your task is to generate a personalized response to the user’s question. Intent refers to the user’s underlying goal, need, or reason for asking the question. Given the question and the user intent, generate a personalized response that directly addresses the user’s needs. Think step by step. Now, answer the following question: \textbf{Question}. User intent: \textbf{intent}.
\vspace{0.3cm}
\hrule
\end{adjustwidth}
\medskip

\section{An Analysis on Base vs. Instruct LLMs.}
Across model families, instruction-tuned variants show only a modest advantage over their corresponding base models in final personalized answer generation. Under IAP, the instruction-tuned versions of Qwen2.5-7B, Gemma3-12B, and Mistral-Small-24B generally achieve slightly higher macro scores, but the overall gap remains small, averaging about 2\% across models. This suggests that although instruction tuning provides a somewhat stronger starting point for learning to understand user intent and translate it into personalized answers, it does not lead to a large difference in final performance once both model types are trained within the same framework. More broadly, this pattern indicates that while general post-training can accelerate early learning in intent-aware reasoning and personalized generation, reinforcement learning can gradually narrow the gap, enabling base models to achieve comparable performance over time.

As shown in Figure~\ref{fig:three_plots}, this pattern is also reflected in the training dynamics. Instruction-tuned models converge faster and begin with stronger initial reward, suggesting that they are better prepared from the start to infer the intent behind a question and generate responses aligned with user-specific needs. Similar differences appear in the generation statistics. For intent length, instruction-tuned models show a sharp early decrease from longer initial intent representations before stabilizing, which suggests that they quickly learn to express user intent in a more concise and task-relevant way. Base models, in contrast, begin with shorter intent representations and adjust more gradually toward similar stable values, indicating a slower process of learning how to capture the user’s underlying goal. A related trend appears in response length: both model types follow a dip-then-rise pattern, but instruction-tuned models recover faster and stabilize earlier, suggesting that they more quickly learn to translate inferred intent into well-formed personalized answers. Overall, these results suggest that instruction tuning mainly improves the speed and efficiency with which models learn to understand user intent and produce personalized responses, while the final gap becomes relatively small once both model types are trained under IAP.
\begin{table*}[t]
\centering
\small
\setlength{\tabcolsep}{7pt}
\renewcommand{\arraystretch}{1.15}

\begin{tabular}{l|cccc|c}
\hline
\textbf{Model} & \textbf{Group Size} & \textbf{Art} & \textbf{Lifestyle} & \textbf{Society} & \textbf{Avg. (macro)} \\
\hline

\multirow{3}{*}{Qwen2.5-7B-it}
& 3 & 0.3219 & 0.4439 & 0.5122 & 0.4260 \\
& 5 & \textbf{0.3533} & \textbf{0.4718} & \textbf{0.5240} & \textbf{0.4497} \\
& 7 & 0.3123 & 0.4312 & 0.5001 & 0.4145 \\
\hline

\multirow{3}{*}{Qwen2.5-7B}
& 3 & 0.3120 & 0.4329 & 0.4429 & 0.3959 \\
& 5 & \textbf{0.3629} & \textbf{0.4889} & \textbf{0.5178} & \textbf{0.4565} \\
& 7 & 0.3213 & 0.4636 & 0.5038 & 0.4295 \\
\hline

\multirow{3}{*}{Gemma3-12B-it}
& 3 & 0.3918 & 0.5533 & 0.6216 & 0.5222 \\
& 5 & \textbf{0.4769} & 0.5934 & \textbf{0.6665} & \textbf{0.5789} \\
& 7 & 0.4593 & \textbf{0.6037} & 0.6488 & 0.5706 \\
\hline

\multirow{3}{*}{Gemma3-12B}
& 3 & 0.4538 & 0.6037 & 0.6069 & 0.5548 \\
& 5 & \textbf{0.4912} & \textbf{0.6218} & \textbf{0.6430} & \textbf{0.5853} \\
& 7 & 0.4759 & 0.5977 & 0.6347 & 0.5694 \\
\hline

\multirow{3}{*}{\shortstack{Mistral-Small-24B-it}}
& 3 & 0.4638 & 0.6439 & 0.6012 & 0.5696 \\
& 5 & \textbf{0.4811} & \textbf{0.6505} & 0.6536 & \textbf{0.5950} \\
& 7 & 0.4420 & 0.6328 & \textbf{0.6612} & 0.5786 \\
\hline

\multirow{3}{*}{\shortstack{Mistral-Small-24B}}
& 3 & 0.4533 & 0.6433 & 0.6323 & 0.5763 \\
& 5 & \textbf{0.4918} & \textbf{0.6649} & \textbf{0.6628} & \textbf{0.6065} \\
& 7 & 0.4655 & 0.6122 & 0.6302 & 0.5693 \\
\hline

\end{tabular}
\caption{IAP performance for different group size. The best result within each model block for each column is bolded.}
\label{tab:group}
\end{table*}
\section{Group Size of IAP}\label{group}

Table~\ref{tab:group} reports the effect of group size on IAP performance across benchmarks. Overall, a group size of 5 yields the strongest average macro performance for all evaluated models, indicating that it provides the most effective balance for policy optimization. In contrast, group size 3 consistently underperforms, suggesting that too few sampled responses limit the quality of relative reward comparison within each group. Group size 7 also fails to improve over 5 in most cases, and often leads to slightly lower scores, implying that increasing the group size beyond a moderate level may introduce additional variance or less informative comparisons~\citep{jin2025searchr}.

This pattern is consistent across both instruction-tuned and base variants. For Qwen2.5-7B-it and Qwen2.5-7B, the average macro score peaks at group size 5, with clear improvements over both 3 and 7. The same trend holds for Gemma3-12B and Gemma3-12B-it, where group size 5 again produces the highest overall performance, although Gemma3-12B-it shows a small exception in the Lifestyle domain, where group size 7 is slightly better. Similarly, both Mistral-Small-24B variants achieve their best average results with group size 5, even though group size 7 gives the highest Society score for the instruction-tuned model.

Taken together, these results suggest that a moderate group size is most effective for IAP. A group size of 5 likely offers enough response diversity to support meaningful relative ranking during optimization, while avoiding the noise and instability that may arise with larger groups. Based on these findings, we use group size 5 in the main experiments.

\end{document}